\documentclass[11pt,letterpaper]{article}
\pdfoutput=1
\usepackage[letterpaper]{geometry}
\usepackage{acl2012}
\usepackage{times}
\usepackage{latexsym}
\usepackage{subfig}
\usepackage{color}
\usepackage{graphicx}
\usepackage{array,multirow,multicol}
\usepackage{epstopdf}
\usepackage{amssymb}
\usepackage{latexsym}
\usepackage{url}
\usepackage{amsmath}
\usepackage{enumitem}
\usepackage{tabularx}
\usepackage{booktabs}
\usepackage{adjustbox}
\usepackage[framemethod=TikZ]{mdframed}
\usepackage{setspace}

\makeatletter
\newcommand{\@BIBLABEL}{\@emptybiblabel}
\newcommand{\@emptybiblabel}[1]{}
\makeatother

\graphicspath{ {./figures/} }

\definecolor{light-gray}{gray}{0.985}
\definecolor{pos}{RGB}{68, 116, 163}
\definecolor{neg}{RGB}{163, 68, 69}

\DeclareMathOperator\softmax{softmax}
\DeclareMathOperator\pol{polarity}
\DeclareMathOperator\gpol{gated-polarity}

\newcommand{\mystar}{{\fontfamily{lmr}\selectfont$\star$}}
\newcommand\doubleRule{\toprule\toprule}
\newcolumntype{R}{>{\raggedleft\arraybackslash}X}
\mdfsetup{
    roundcorner=1.5pt,
    backgroundcolor=light-gray,
    linewidth=1pt,
    frametitlerule=true, 
}

\title{Multiple Instance Learning Networks for Fine-Grained Sentiment Analysis}

 \author{Stefanos Angelidis \textnormal{and} Mirella Lapata \\
   Institute for Language, Cognition and Computation \\
   School of Informatics, University of Edinburgh \\
   10 Crichton Street, Edinburgh EH8 9AB \\
   {\tt s.angelidis@ed.ac.uk, mlap@inf.ed.ac.uk} \\\\
}

\date{}

\begin{document}

\maketitle

\begin{abstract}
  We consider the task of fine-grained sentiment analysis
  from the perspective of multiple instance learning (MIL). Our neural 
  model is trained on document sentiment labels, and learns to predict the
  sentiment of text segments, i.e. sentences or elementary discourse
  units (EDUs), without segment-level supervision. We introduce an
  attention-based polarity scoring method for identifying positive 
  and negative text snippets and a new dataset which we call \textsc{SpoT} 
  (as shorthand for \textbf{S}egment-level \textbf{PO}lari\textbf{T}y 
  annotations) for evaluating MIL-style sentiment models like ours. 
  Experimental results demonstrate superior performance against multiple 
  baselines, whereas a judgement elicitation study shows that EDU-level opinion 
  extraction produces more informative summaries than sentence-based
  alternatives.

\end{abstract}

\section{Introduction}
Sentiment analysis has become a fundamental area of research in
Natural Language Processing thanks to the proliferation of
user-generated content in the form of online reviews, blogs, internet
forums, and social media.  A plethora of methods have been proposed in
the literature that attempt to distill sentiment information from
text, allowing users and service providers to make
opinion-driven decisions.

\begin{figure}[t]
	\begin{mdframed}[frametitle={\footnotesize
\textnormal{[Rating: {\Large \mystar \mystar}] I had a very mixed experience at
The Stand. The burger and fries were good. The chocolate shake was divine: rich 
and creamy. The drive-thru was horrible. It took us at least 30 minutes to order 
when there were only four cars in front of us. We complained about 
the wait and got a half--hearted apology. I would go back because the 
food is good, but my only hesitation is the wait.}}]

		\rotatebox[origin=c]{90}{Summary}\hspace{-1mm}
	    {\centering
\adjustbox{minipage=\columnwidth,margin=0em,width=\columnwidth,center,bgcolor=light-gray}{%
	        \small
	    	\textcolor{pos}{~~~~+ The burger and fries were good}\\
	    	\textcolor{pos}{~~~~+ The chocolate shake was divine}\\
		    \textcolor{pos}{~~~~+ I would go back because the food is good}\\
		    \textcolor{neg}{~~~~-- The drive-thru was horrible}\\
		    \textcolor{neg}{~~~~-- It took us at least 30 minutes to order}%
		}}
	\end{mdframed}
    \vspace{-3mm}
    \caption{An EDU-based summary of a 2-out-of-5 star review with positive 
    and negative snippets.}
   \vspace{-4mm}
    \label{fig:example-summ}
\end{figure}

The success of neural networks in a variety of applications
\cite{bahdanau2014neural,le2014distributed,socher2013recursive} and
the availability of large amounts of labeled data have led to an
increased focus on sentiment classification. Supervised models are typically
trained on documents
\cite{johnson2015effective,johnson2015semi,tang2015document,yang2016hierarchical},
sentences \cite{kim2014convolutional}, or phrases
\cite{socher2011semi,socher2013recursive} annotated with sentiment
labels and used to predict sentiment in unseen texts.  Coarse-grained
document-level annotations are relatively easy to obtain due to the
widespread use of opinion grading interfaces (e.g.,~star ratings
accompanying reviews). In contrast, the acquisition of sentence- or
phrase-level sentiment labels remains a laborious and expensive
endeavor despite its relevance to various opinion mining applications,
e.g., detecting or summarizing consumer opinions in online product
reviews. The usefulness of \mbox{finer-grained} sentiment analysis is illustrated
in the example of Figure~\ref{fig:example-summ}, where snippets 
of opposing polarities are extracted from a 2-star restaurant review. Although, 
as a whole, the review conveys negative sentiment, aspects of the reviewer's
experience were clearly positive. This goes largely unnoticed when focusing 
solely on the review's overall rating.

In this work, we consider the problem of segment-level sentiment
analysis from the perspective of \textit{Multiple Instance Learning} (MIL;
Keeler, 1991). Instead of learning from individually labeled segments,
our model only requires document-level supervision and learns to
introspectively judge the sentiment of constituent segments. Beyond
showing how to utilize document collections of rated reviews to train
\mbox{fine-grained} sentiment predictors, we also investigate the
granularity of the extracted segments. Previous research
\cite{tang2015document,yang2016hierarchical,cheng2016neural,Nallapati:ea:17}
has predominantly viewed documents as sequences of sentences. Inspired
by recent work in summarization \cite{li2016role} and sentiment
classification \cite{bhatia2015better}, we also represent documents
via Rhetorical Structure Theory's \cite{mann1988rhetorical}
\textit{Elementary Discourse Units} (EDUs). Although definitions for
EDUs vary in the literature, we follow standard practice and take the
elementary units of discourse to be clauses
\cite{carlson2003building}.  We employ a state-of-the-art discourse
parser \cite{feng2012text} to identify them.

Our contributions in this work are three-fold: a novel multiple
instance learning neural model which utilizes document-level sentiment
supervision to judge the polarity of its constituent segments; the
creation of \textsc{SpoT}, a publicly available dataset which contains
\textbf{S}egment-level \textbf{PO}lari\textbf{T}y annotations (for
sentences and EDUs) and can be used for the evaluation of MIL-style
models like ours; and the empirical finding (through automatic and
human-based evaluation) that neural multiple instance learning is
superior to more conventional neural architectures and other baselines
on detecting segment sentiment and extracting informative opinions in
reviews.\footnote{Our code and \textsc{SpoT} dataset are publicly available at:
  \url{https://github.com/stangelid/milnet-sent}}

\section{Background}
\label{sec:background}

Our work lies at the intersection of multiple research areas,
including sentiment classification, opinion mining and multiple
instance learning. We review related work in these areas below.

\paragraph{Sentiment Classification} 
Sentiment classification is one of the most popular
tasks in sentiment analysis. Early work focused on unsupervised
methods and the creation of sentiment lexicons
\cite{turney2002thumbs,hu2004mining,wiebe2005annotating,baccianella10sentiwordnet}
based on which the overall polarity of a text can be computed
(e,g.,~by aggregating the sentiment scores of constituent words). More
recently, \newcite{taboada2011lexicon} introduced \mbox{SO-CAL}, a
\mbox{state-of-the-art} method that combines a rich sentiment lexicon
with carefully defined rules over syntax trees to predict sentence
sentiment. 

Supervised learning techniques have subsequently dominated the
literature
\cite{pang2002thumbs,pang2005seeing,qu2010bag,xia2010exploring,wang2012baselines,le2014distributed}
thanks to user-generated sentiment labels or large-scale
crowd-sourcing efforts \cite{socher2013recursive}.  Neural network
models in particular have achieved state-of-the-art performance on
various sentiment classification tasks due to their ability to
alleviate feature engineering.  \newcite{kim2014convolutional}
introduced a very successful CNN architecture for sentence-level
classification, whereas other work
\cite{socher2011semi,socher2013recursive} uses recursive neural
networks to learn sentiment for segments of varying granularity
(i.e.,~words, phrases, and sentences). We describe Kim's
\shortcite{kim2014convolutional} approach in more detail as it is also
used as part of our model.

Let~$\mathrm{x}_i$ denote a $k$-dimensional word embedding of
the~$i$-th word in text segment~$s$ of length~$n$. The segment's input
representation is the concatenation of word embeddings $\mathrm{x}_1,
\dots, \mathrm{x}_n$, resulting in word matrix~$X$. Let~$X_{i:i+j}$
refer to the concatenation of embeddings $\mathrm{x}_i, \dots,
\mathrm{x}_{i+j}$. A convolution filter $W \in \mathbb{R}^{lk}$,
applied to a window of $l$~words, produces a new feature $c_i =
\mathrm{ReLU}(W \circ X_{i:i+l} + b)$, where ReLU is the
\textit{Rectified Linear Unit} non-linearity, `$\circ$' denotes the
entrywise product followed by a sum over all elements and $b \in
\mathbb{R}$ is a bias term. Applying the same filter to every possible
window of word vectors in the segment, produces a feature map
\mbox{$\mathrm{c} = [c_1, c_2, \dots, c_{n-l+1}]$}.  Multiple feature
maps for varied window sizes are applied, resulting in a
\mbox{fixed-size} segment representation $\mathrm{v}$ via
max-over-time pooling.  We will refer to the application of
convolution to an input word matrix~$X$, as $\mathrm{CNN}(X)$. A final
sentiment prediction is produced using a softmax classifier and the
model is trained via back-propagation using sentence-level sentiment
labels.

The availability of large-scale datasets
\cite{wujointly,tang2015document} has also led to the development of
document-level sentiment classifiers which exploit hierarchical neural
representations.  These are obtained by first building representations
of sentences and aggregating those into a document feature vector
\cite{tang2015document}. \newcite{yang2016hierarchical} further
acknowledge that words and sentences are deferentially important in
different contexts. They present a model which learns to attend
\cite{bahdanau2014neural} to individual text parts when constructing
document representations. We describe such an architecture in more detail
as we use it as a point of comparison with our own model.

\begin{figure*}[t]
	\centering
	\includegraphics[width=.9\textwidth]{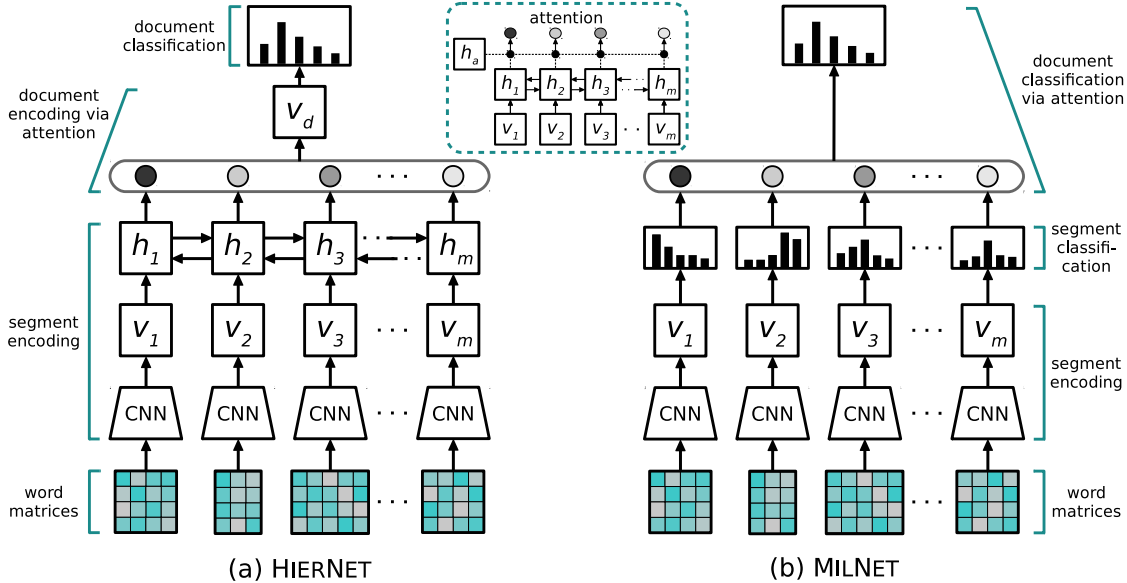}
	\caption{A \textit{Hierarchical Network} (\textsc{HierNet}) for 
	document-level sentiment classification and our proposed \textit{Multiple 
	Instance Learning Network} (\textsc{MilNet}). The models use the same 
	attention mechanism to combine segment vectors and predictions respectively.}
	\label{fig:networks}
\end{figure*}

Given document $d$ comprising segments $(s_1, \dots, s_m)$, a
\textit{Hierarchical Network} with attention (henceforth \textsc{HierNet}; 
based on Yang et al., 2016) produces segment representations $(\mathrm{v}_1, \dots,
\mathrm{v}_m)$ which are subsequently fed into a bidirectional GRU
module \cite{bahdanau2014neural}, whose resulting hidden vectors
$(\mathrm{h}_1, \dots, \mathrm{h}_m)$ are used to produce attention
weights $(a_1, \dots, a_m)$ (see Section~\ref{sec:milmodels} for more
details on the attention mechanism).  A document is represented as the
weighted average of the segments' hidden vectors~$\mathrm{v}_d = \sum_i a_i
\mathrm{h}_i$. A final sentiment prediction is obtained using a softmax
classifier and the model is trained via back-propagation using
document-level sentiment labels. The architecture is illustrated in
Figure~\ref{fig:networks}(a). In their proposed model, 
\newcite{yang2016hierarchical} use bidirectional GRU modules to represent 
segments as well as documents, whereas we use a more efficient CNN encoder 
to compose words into segment vectors\footnote{When applied 
to the YELP'13 and IMDB document classification datasets, the use of CNNs 
results in a relative performance decrease of $<2\%$ compared Yang 
et al's model (2016).} (i.e.,~$\mathrm{v}_i = \mathrm{CNN}(X_i)$). Note that 
models like \textsc{HierNet} do not naturally predict sentiment for individual
segments; we discuss how they can be used for segment-level opinion
extraction in Section~\ref{sec:comparison}.

Our own work draws inspiration from representation learning
\cite{tang2015document,kim2014convolutional}, especially the idea that
not all parts of a document convey sentiment-worthy clues
\cite{yang2016hierarchical}. Our model departs from previous
approaches in that it provides a natural way of predicting the
polarity of \emph{individual} text segments without requiring
segment-level annotations. Moreover, our attention mechanism
directly facilitates opinion detection rather than simply
aggregating sentence representations into a single document vector.

\paragraph{Opinion Mining} A standard setting for opinion
mining and summarization
\cite{lerman-blairgoldensohn-mcdonald:2009:EACL,Carenini:ea:2006,ganesan-zhai-han:2010:PAPERS,difabbrizio-stent-gaizauskas:2014:W14-44,gerani-EtAl:2014:EMNLP2014}
assumes a set of documents that contain opinions about some entity of
interest (e.g.,~camera). The goal of the system is to generate a
summary that is representative of the average opinion and speaks to
its important aspects (e.g.,~picture quality, battery life, value).
Output summaries can be extractive
\cite{lerman-blairgoldensohn-mcdonald:2009:EACL} or abstractive
\cite{gerani-EtAl:2014:EMNLP2014,difabbrizio-stent-gaizauskas:2014:W14-44}
and the underlying systems exhibit varying degrees of linguistic
sophistication from identifying aspects
\cite{lerman-blairgoldensohn-mcdonald:2009:EACL} to using RST-style
discourse analysis, and manually defined templates
\cite{gerani-EtAl:2014:EMNLP2014,difabbrizio-stent-gaizauskas:2014:W14-44}.

Our proposed method departs from previous work in that it focuses on
detecting opinions in individual documents. Given a review, we predict the
polarity of every segment, allowing for the extraction of sentiment-heavy
opinions. We explore the usefulness of EDU segmentation inspired
by \newcite{li2016role}, who show that EDU-based summaries align with 
near-extractive summaries constructed by news editors. Importantly, 
our model is trained in a weakly-supervised fashion on large scale document 
classification datasets without recourse to fine-grained labels or 
gold-standard opinion summaries.

\paragraph{Multiple Instance Learning} Our models adopt a \textit{Multiple
Instance Learning} (MIL) framework. MIL deals with problems where labels are 
associated with groups of instances or \textit{bags} (documents in our case), 
while instance labels (segment-level polarities) are unobserved. An aggregation 
function is used to combine instance predictions and assign labels on the bag 
level. The goal is either to label bags 
\cite{keeler1991integrated,DIETTERICH199731,maron1998multiple}
or to simultaneously infer bag and instance labels 
\cite{zhou2009multi,wei2014scalable,kotzias2015group}. We view 
segment-level sentiment analysis as an instantiation of the latter variant.

Initial MIL efforts for binary classification made the strong
assumption that a bag is negative only if all of its instances are
negative, and positive otherwise 
\cite{DIETTERICH199731,maron1998multiple,zhang2002content,andrews2003support,carbonetto2008learning}.
Subsequent work relaxed this assumption, allowing for prediction
combinations better suited to the tasks at
hand. \newcite{weidmann2003twolevel} introduced a generalized MIL
framework, where a combination of instance types is required to assign
a bag label. \newcite{zhou2009multi} used graph kernels to aggregate
predictions, exploiting relations between instances in object and text
categorization. \newcite{xu04logisticregression} proposed a
\mbox{multiple-instance} logistic regression classifier where instance
predictions were simply averaged, assuming equal and independent
contribution toward bag classification. More recently, 
\newcite{kotzias2015group} used sentence vectors obtained by a pre-trained 
hierarchical CNN \cite{denil2014extraction} as features under an unweighted 
average MIL objective. Prediction averaging was further extended by
%\newcite{pappas2014explaining,pappas2017explicit}, 
Pappas and Popescu-Belis (2014; 2017),
who used a weighted summation of predictions, an idea which we also adopt
in our work. \nocite{pappas2017explicit}

Applications of MIL are many and varied. MIL was first explored by
\newcite{keeler1991integrated} for recognizing handwritten post codes,
where the position and value of individual digits was unknown. MIL
techniques have since been applied to drug activity prediction
\cite{DIETTERICH199731}, image retrieval
\cite{maron1998multiple,zhang2002content}, object detection
\cite{zhang2006multiple,carbonetto2008learning,cour2011learning}, text
classification \cite{andrews2003support}, image captioning
\cite{wu2015deep}, paraphrase detection \cite{xu2014extracting}, and
information extraction \cite{hoffmann2011knowledge}. 

When applied to sentiment analysis, MIL takes advantage of supervision signals 
on the document level in order to train segment-level sentiment predictors.
Although their work is not couched in the framework of MIL,
\newcite{tackstrom2011discovering} show how sentence sentiment labels
can be learned as latent variables from document-level annotations
using hidden conditional random fields. \newcite{pappas2014explaining} 
use a multiple instance regression model to assign sentiment scores to specific 
aspects of products. The Group-Instance Cost Function (GICF), proposed by 
\newcite{kotzias2015group}, averages sentence sentiment predictions during
trainng, while ensuring that similar sentences receive similar 
polarity labels. Their work uses a pre-trained hierarchical CNN to
obtain sentence embeddings, but is not trainable end-to-end, in contrast with
our proposed network.  Additionally, none of the aforementioned
efforts explicitly evaluate opinion extraction quality.

%\paragraph{Convolutional Neural Networks}

%\noindent This architecture is similar to the hierarchical attention
%networks (HAN) of \newcite{yang2016hierarchical}. Their model uses the
%same GRU-based attention mechanism on both word and segment levels,
%whereas we use a more efficient CNN encoder to compose word
%vectors.\footnote{When applied to document classification (which is
%  not the focus of this work), \mbox{\textsc{HierNet}} achieves 66.8\%
%  and 47.5\% accuracy on YELP'13 and IMDB, respectively, compared to
%  HAN's 68.2\% and 49.4\%.}

\section{Methodology}
In this section we describe how multiple instance learning can be used
to address some of the drawbacks seen in previous approaches, namely
the need for expert knowledge in lexicon-based sentiment analysis
\cite{taboada2011lexicon}, expensive fine-grained annotation on the
segment level \cite{kim2014convolutional,socher2013recursive} or the
inability to naturally predict segment sentiment
\cite{yang2016hierarchical}.

\subsection{Problem Formulation}
\label{sec:problem}

Under multiple instance learning (MIL), a dataset~$D$ is a collection
of labeled \textit{bags}, each of which is a group of unlabeled
\textit{instances}. Specifically, each document~$d$ is a sequence
(bag) of segments (instances). This sequence \mbox{$ d = (s_1, s_2,
  \dots, s_m)$} is obtained from a document segmentation policy (see
Section~\ref{sec:segm} for details).  A discrete sentiment label $y_d \in [1,C]$ 
is associated with each document, where the labelset is
ordered and classes~$1$ and $C$~correspond to maximally negative and
maximally positive sentiment. It is assumed that $y_d$ is an unknown
function of the unobserved segment-level labels:
\begin{equation}
\label{eq:mil-labels}
y_d = f(y_{1}, y_{2}, \dots, y_{m})
\end{equation}

\noindent Probabilistic sentiment classifiers will produce
document-level predictions~$\hat{y}_d$ by selecting the most probable
class according to class distribution \mbox{$\mathrm{p}_d = \langle
  p_d^{(1)},\dots,p_d^{(C)} \rangle$}. In a non-MIL framework a
classifier would learn to predict the document's sentiment by directly
conditioning on its segments' feature representations or their aggregate:
\vspace{-1mm}
\begin{equation}
\label{eq:nonmil-preds}
\mathrm{p}_d = \hat{f}_\theta(\mathrm{v}_1, \mathrm{v}_2, \dots, \mathrm{v}_m)
\vspace{-1mm}
\end{equation}

\noindent In contrast, a MIL classifier will produce a class 
distribution $p_i$ for each segment and additionally learn to 
combine these into a document-level prediction:
\vspace{-2mm}
\begin{align}
\mathrm{p}_{i\,} &= \hat{g}_{\theta_s}(\mathrm{v}_i)\,, \label{eq:mil-spreds} \\
\mathrm{p}_{d} &= \hat{f}_{\theta_d}(\mathrm{p}_{1}, \mathrm{p}_{2}, \dots, \mathrm{p}_{m})\,. \label{eq:mil-dpreds}
\vspace{-4mm}
\end{align}
In this work, $\hat{g}$ and $\hat{f}$ are defined using a single
neural network, described below.

\subsection{Multiple Instance Learning Network}
\label{sec:milmodels}

Hierarchical neural models like \textsc{HierNet} have
been used to predict document-level polarity by first encoding
sentences and then combining these representations into
a document vector. Hierarchical vector composition produces powerful
sentiment predictors, but lacks the ability to introspectively judge
the polarity of individual segments.

Our \textit{Multiple Instance Learning Network} (henceforth \textsc{MilNet}) 
is based on the following intuitive assumptions about opinionated text. Each
segment conveys a degree of sentiment polarity, ranging from very
negative to very positive. Additionally, segments have varying degrees
of importance, in relation to the overall opinion of the author. The
overarching polarity of a text is an aggregation of segment
polarities, weighted by their importance. Thus, our model attempts to
predict the polarity of segments and decides which parts of the
document are good indicators of its overall sentiment, allowing for
the detection of \mbox{sentiment-heavy} opinions. An illustration of
\textsc{MilNet} is shown in Figure~\ref{fig:networks}(b); the model
consists of three components: a CNN segment encoder, a softmax segment
classifier and an attention-based prediction weighting module.

\begin{figure*}[t!]
	\centering
	\begin{minipage}{\textwidth}
	$\quad\;\;\;\,\;\,\,$The starters were quite bland.
	$\quad\,\;\;\;$I didn't enjoy most of them,
	$\quad\;\;\;\;\,$but the burger was brilliant!
	\end{minipage}
	\begin{minipage}{\textwidth}
	\includegraphics[width=\textwidth]{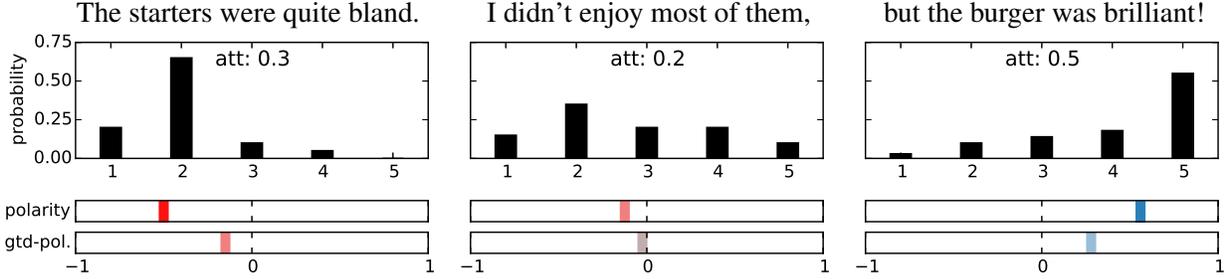}
	\end{minipage}
\vspace{-2ex}
	\caption{Polarity scores (bottom) obtained from class
          probability distributions for three EDUs (top) extracted
          from a restaurant review. Attention weights (top) 
          are used to fine-tune the obtained polarities.}
	\label{fig:polarity}
\vspace{-1ex}
\end{figure*}

\paragraph{Segment Encoding}
An encoding $\mathrm{v}_i = \mathrm{CNN}(X_i)$ is produced for each
segment, using the CNN architecture described in
Section~\ref{sec:background}.

\paragraph{Segment Classification} 
Obtaining a separate representation
$\mathrm{v}_i$ for every segment in a document allows us to produce individual
segment sentiment predictions 
\mbox{$\mathrm{p}_i = \langle p_i^{(1)},\dots,p_i^{(C)} \rangle$}.
This is achieved using a softmax classifier:
\begin{equation}
\mathrm{p}_i = \softmax(W_{c}\mathrm{v}_i + b_{c})\,,
\end{equation}
\noindent where $W_c$ and $b_c$ are the classifier's parameters, 
shared across all segments. Individual distributions $p_i$
are shown in Figure \ref{fig:networks}(b) as small bar-charts.

\paragraph{Document Classification} In the simplest case,
document-level predictions can be produced by taking the average of
segment class distributions: $p_d^{(c)} =\ ^1/_m \sum_i p_i^{(c)}\,,\,
c \in [1,C]$.  This is, however, a crude way of combining segment
sentiment, as not all parts of a document convey important sentiment
clues. We opt for a segment attention mechanism which rewards text
units that are more likely to be good sentiment predictors.

Our attention mechanism is based on a bidirectional GRU component
\cite{bahdanau2014neural} and inspired by
\newcite{yang2016hierarchical}. However, in contrast to their work,
where attention is used to combine sentence representations into a
single document vector, we utilize a similar technique to aggregate
individual sentiment \emph{predictions}.

We first use separate GRU modules to produce forward and backward 
hidden vectors, which are then concatenated:
\begin{align}
\overrightarrow{\mathrm{h}}_i &= \overrightarrow{\mathrm{GRU}}(\mathrm{v}_i), \label{eq:gru1}\ \\
\overleftarrow{\mathrm{h}}_i &= \overleftarrow{\mathrm{GRU}}(\mathrm{v}_i), \label{eq:gru2}\ \\
\mathrm{h}_i &= [\overrightarrow{\mathrm{h}}_i, \overleftarrow{\mathrm{h}}_i], \; i \in [1,m]\,. \label{eq:gru3}\ 
\end{align}
\vspace{-7mm}

\noindent The importance of each segment is measured with the aid of a 
vector $\mathrm{h}_a$, as follows:
\begin{align}
\mathrm{h}_i' &= \tanh(W_a\mathrm{h}_i + \mathrm{b}_a)\,, \label{eq:att1}\ \\
a_i  &= \frac{exp(\mathrm{h}_i'^{\mathsf{T}}\mathrm{h}_a)}{\sum_i exp(\mathrm{h}_i'^{\mathsf{T}}\mathrm{h}_a)}\,, \label{eq:att2}\
\end{align}
\noindent where Equation~(\ref{eq:att1}) defines a one-layer MLP that
produces an attention vector for the $i$-th segment.  Attention
weights $a_i$ are computed as the normalized similarity of each $\mathrm{h}_i'$
with $\mathrm{h}_a$. Vector $\mathrm{h}_a$, which is randomly initialized and 
learned during training, can be thought of as a trained \textit{key}, 
able to recognize sentiment-heavy segments. The attention mechanism is
depicted in the dashed box of Figure~\ref{fig:networks}, with
attention weights shown as shaded circles.

Finally, we obtain a document-level distribution over sentiment labels
as the weighted sum of segment distributions (see top of Figure
\ref{fig:networks}(b)):
\begin{equation}
\label{eq:sum}
p_d^{(c)} = \sum_i a_i p_i^{(c)}\,, \; c \in [1,C]\,.
\end{equation}
\vspace{-7mm}

\paragraph{Training} The model is trained end-to-end on documents with
user-generated sentiment labels. We use the negative log likelihood of
the document-level prediction as an objective function:
\begin{equation}
L = -\sum_d\log p_d^{(y_d)}
\end{equation}

\section{Polarity-based Opinion Extraction}
\label{sec:polarity}

After training, our model can produce segment-level sentiment
predictions for unseen texts in the form of class probability
distributions. A direct application of our method is opinion
extraction, where highly positive and negative snippets are selected
from the original document, producing extractive sentiment summaries,
as described below.

\paragraph{Polarity Scoring} 
In order to extract opinion summaries, we need to rank segments according
to their sentiment polarity. We introduce a method that takes our model's 
confidence in the prediction into account, by reducing each segment's class
probability distribution~$\mathrm{p}_i$ to a single real-valued polarity score. 
To achieve this, we first define a real-valued \textit{class weight} vector 
\mbox{$\mathrm{w} = \langle w^{(1)}, \dots, w^{(C)} \, | \, w^{(c)} \in [-1,1]\rangle$} 
that assigns uniformly-spaced weights to the ordered labelset, such that 
\mbox{$w^{(c+1)}-w^{(c)} =\frac{2}{C-1}$}. For example, in a 5-class scenario, 
the class weight vector would be $\mathrm{w} = \langle-1, -0.5, 0, 0.5, 1\rangle$. 
We compute the polarity score of a segment as the dot-product of the probability
distribution~$\mathrm{p}_i$ with vector~$\mathrm{w}$:
\begin{equation}
\pol(s_i) = \sum_c p_i^{(c)} w^{(c)} \;\; \in [-1,1]
\end{equation}

\paragraph{Gated Polarity}
As a way of increasing the effectiveness of our method, we 
introduce a \textit{gated} extension that uses the attention
mechanism of our model to further differentiate between segments that 
carry significant sentiment cues and those that do not:
\begin{equation}
\gpol(s_i) = a_i \cdot \pol(s_i)\,,
\label{eq:gating}
\end{equation}
\noindent where $a_i$ is the attention weight assigned to the
\mbox{$i$-th} segment. This forces the polarity scores of segments the
model does not attend to closer to 0.

An illustration of our polarity scoring function is provided in Figure
\ref{fig:polarity}, where the class predictions (top) of three
restaurant review segments are mapped to their corresponding polarity
scores (bottom). We observe that our method produces the desired
result; segments 1 and 2 convey negative sentiment and
receive negative scores, whereas the third segment is mapped to a
positive score. Although the same discrete class label is assigned 
to the first two, the second segment's score is closer to 0 (neutral) as 
its class probability mass is more evenly distributed.

\paragraph{Segmentation Policies}
\label{sec:segm}

As mentioned earlier, one of the hypotheses investigated in this work
regards the use of subsentential units as the basis of
extraction. Specifically, our model was applied to sentences and
\textit{Elementary Discourse Units} (EDUs), obtained from a Rhetorical
Structure Theory (RST) parser \cite{feng2012text}. According to RST,
documents are first segmented into EDUs corresponding roughly to
independent clauses which are then recursively combined into larger
discourse spans. This results in a tree representation of the
document, where connected nodes are characterized by discourse
relations. We only utilize RST's segmentation, and leave the potential
use of the tree structure to future work.

The example in Figure~\ref{fig:polarity} illustrates why EDU-based
segmentation might be beneficial for opinion extraction.  The
second and third EDUs correspond to the sentence: \textsl{I didn't
  enjoy most of them, but the burger was brilliant}. Taken as a whole,
the sentence conveys mixed sentiment, whereas the EDUs clearly convey
opposing sentiment.

\begin{table}[t]
\centering
\small
\begin{tabularx}{\columnwidth}{lRR}
\toprule
%& \multicolumn{2}{c}{\textbf{$\quad\;$Collections}}\\
& \textbf{Yelp'13} & \textbf{IMDB}\\
\doubleRule
Documents & 335,018 & 348,415\\
Average \#Sentences & 8.90 & 14.02\\
Average \#EDUs & 19.11 & 37.38\\
Average \#Words & 152 & 325\\
Vocabulary Size & 211,245 & 115,831\\
Classes & 1--5 & 1--10\\
\bottomrule
\end{tabularx}
\vspace{-1ex}
\caption{Document-level sentiment classification datasets used to train our models.}
\label{tbl:classdatasets}
\end{table}

\begin{table}[t]
\centering
\small
\begin{tabularx}{\columnwidth}{lXXXX}
\toprule
%& \multicolumn{2}{c}{\textbf{$\quad\;$Collections}}\\
& \multicolumn{2}{c}{\textbf{Yelp'13$_{seg}$}} & \multicolumn{2}{c}{\textbf{IMDB$_{seg}$}}\\
& Sent. & EDUs & Sent. & EDUs\\
\doubleRule
\#Segments & 1,065 & 2,110 & 1,029 & 2,398\\
\#Documents & \multicolumn{2}{c}{100} & \multicolumn{2}{c}{97}\\
Classes & \multicolumn{2}{c}{\{--~,~0~,~+\}} & \multicolumn{2}{c}{\{--~,~0~,~+\}}\\
\bottomrule
\end{tabularx}
\vspace{-1ex}
\caption{\textsc{SpoT} dataset: numbers of documents and segments with
  polarity annotations.}
\label{tbl:segdatasets}
\end{table}

\begin{figure}[t]
	\centering
	\begin{minipage}{\columnwidth}
	\includegraphics[width=\textwidth]{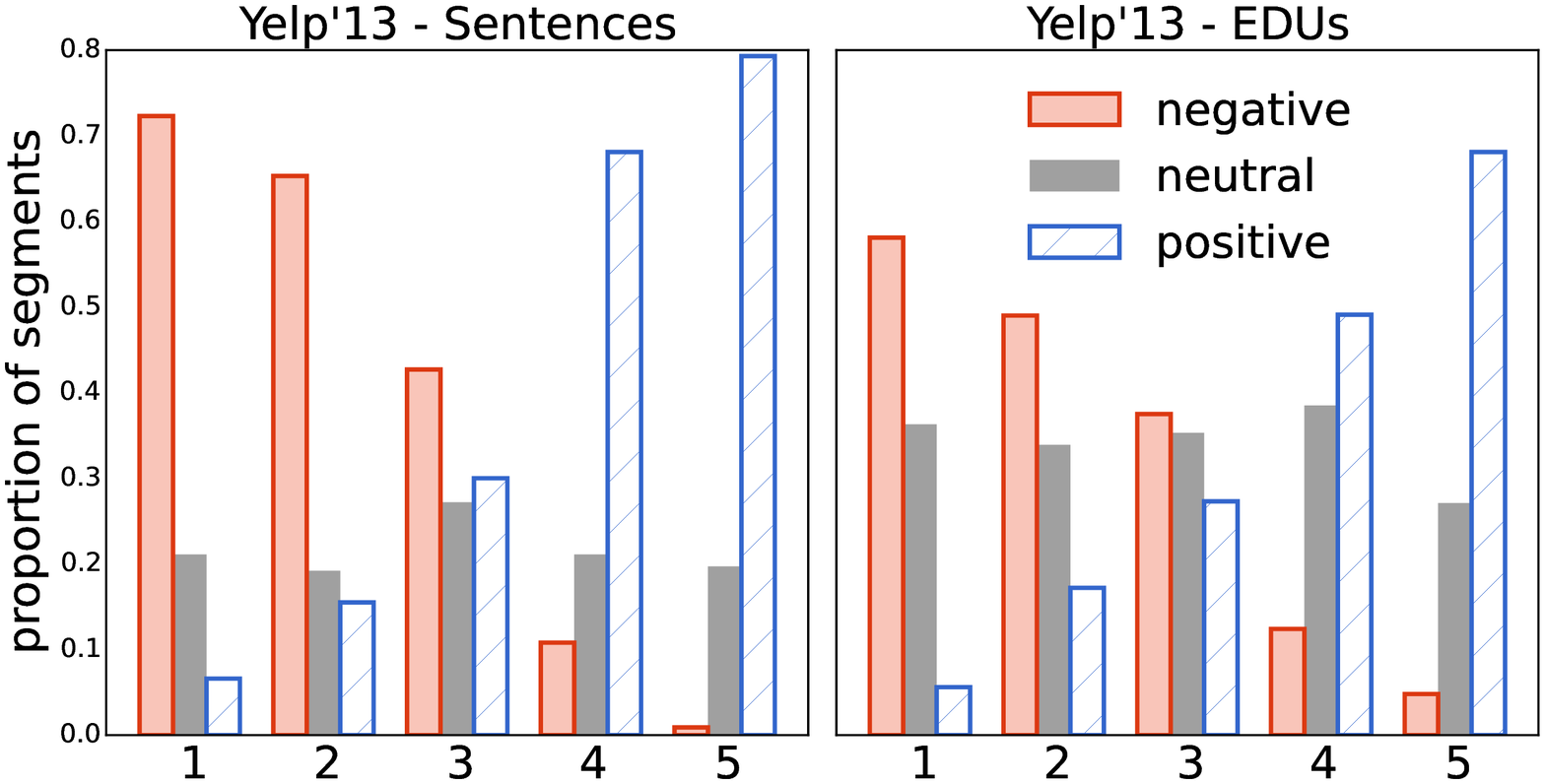}
	\end{minipage}\vspace{1.3mm}
	\begin{minipage}{\columnwidth}
	\includegraphics[width=\textwidth]{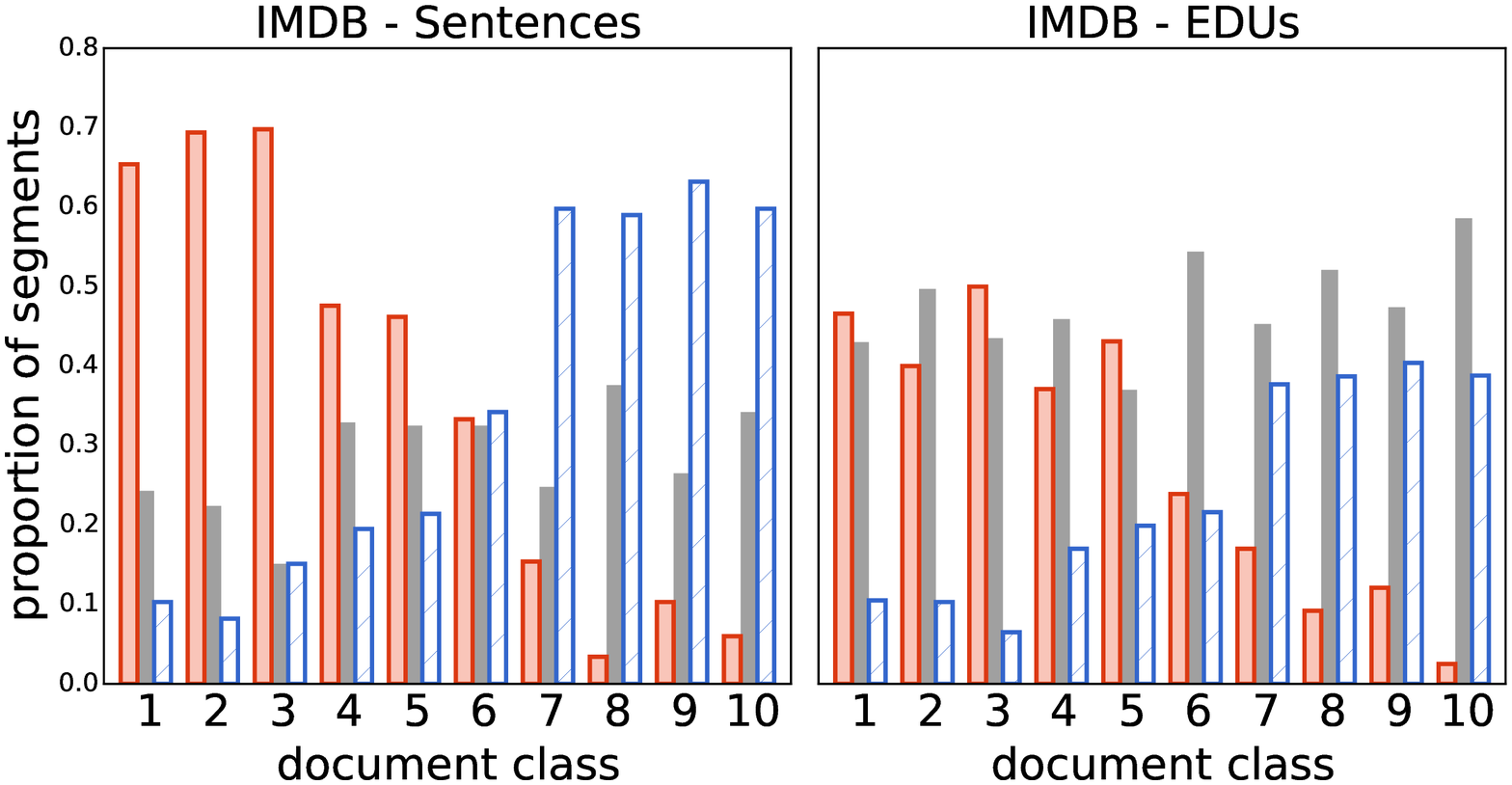}
	\end{minipage}
\vspace{-1ex}
	\caption{Distribution of segment-level labels per document-level 
class on our the \textsc{SpoT} datasets.}
	\label{fig:slabels}
\end{figure}

\section{Experimental Setup}
\label{sec:exp}

In this section we describe the data used to assess the performance of
our model. We also give details on model training and comparison
systems. 

\subsection{Datasets}
\label{sec:data}

Our models were trained on two large-scale sentiment classification
collections. The Yelp'13 corpus was introduced in
\newcite{tang2015document} and contains customer reviews of local
businesses, each associated with human ratings on a scale from 1
(negative) to 5 (positive). The IMDB corpus of movie reviews was
obtained from \newcite{wujointly}; each review is associated with user
ratings ranging from~1 to~10. Both datasets are split into training
(80\%), validation (10\%) and test (10\%) sets. A summary of
statistics for each collection is provided in
Table~\ref{tbl:classdatasets}.

In order to evaluate model performance on the segment level, we
constructed a new dataset named \textsc{SpoT} (as a shorthand for
\textbf{S}egment \textbf{PO}lari\textbf{T}y) by annotating documents
from the Yelp'13 and IMDB collections.
Specifically, we sampled reviews from each collection such that
all document-level classes are represented uniformly, and the document
lengths are representative of the respective corpus.  Documents were
segmented into sentences and EDUs, resulting in two segment-level
datasets per collection. Statistics are summarized in
Table~\ref{tbl:segdatasets}.

Each review was presented to three \textit{Amazon Mechanical Turk} (AMT)
annotators who were asked to judge the sentiment conveyed by each
segment (i.e.,~sentence or EDU) as \textit{negative},
\textit{neutral}, or \textit{positive}. We assigned labels using a
majority vote or a fourth annotator in the rare cases of no agreement
($<5\%$). Figure~\ref{fig:slabels} shows the distribution of segment
labels for each document-level class. As expected, documents with
positive labels contain a larger number of positive segments compared
to documents with negative labels and vice versa. Neutral segments are
distributed in an approximately uniform manner across document
classes. Interestingly, the proportion of neutral EDUs is
significantly higher compared to neutral sentences.  The observation
reinforces our argument in favor of EDU segmentation, as it suggests
that a sentence with positive or negative overall polarity may still
contain neutral EDUs. Discarding neutral EDUs, could therefore lead to
more concise opinion extraction compared to relying on entire sentences.

We further experimented on two collections introduced by
\newcite{kotzias2015group} which also originate from the YELP'13 and
IMDB datasets. Each collection consists of 1,000 randomly sampled
sentences annotated with binary sentiment labels.

\begin{figure*}[t]
	\centering
	\begin{minipage}{\textwidth}
	\includegraphics[width=\textwidth]{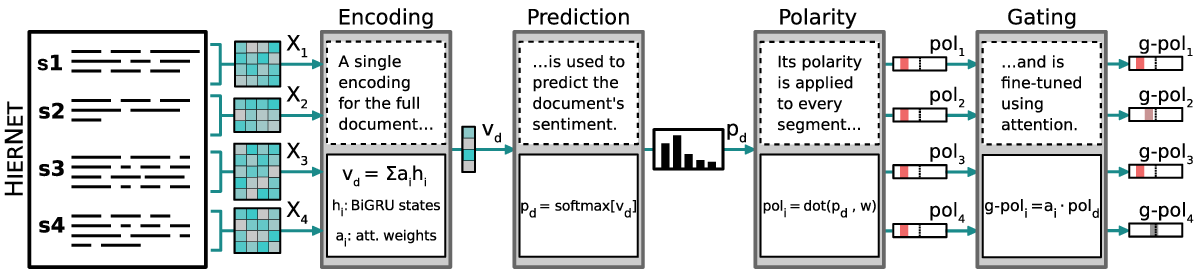}
	\end{minipage}\vspace{1mm}
	\begin{minipage}{\textwidth}
	\includegraphics[width=\textwidth]{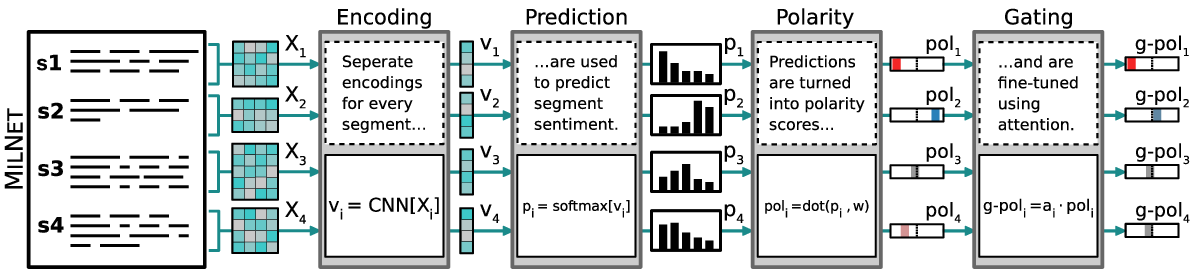}
	\end{minipage}
	\caption{System pipelines for \textsc{HierNet} and \textsc{MilNet}
	showing 4 distinct phases for sentiment analysis.}
	\label{fig:pipeline}
\end{figure*}

\subsection{Model Comparison}
\label{sec:comparison}

On the task of segment classification we compared \textsc{MilNet}, our
multiple instance learning network, against the following methods:
\begin{itemize}[label={},itemindent=-1em,leftmargin=1em,itemsep=0em]
\item \textbf{Majority:} Majority class applied to all instances.
\item \textbf{SO-CAL:} State-of-the-art lexicon-based system that
  classifies segments into positive, neutral, and negative classes
  \cite{taboada2011lexicon}.
\item \textbf{Seg-CNN:} Fully-supervised CNN segment classifier
	trained on \textsc{SpoT}'s labels \cite{kim2014convolutional}.
\item \textbf{GICF:} The Group-Instance Cost Function model
      introduced in \newcite{kotzias2015group}. This is an unweighted
      average prediction aggregation MIL method that uses sentence features
      from a pre-trained convolutional neural model.
\item \textbf{\textsc{HierNet}:} \textsc{HierNet} does
      not explicitly generate individual segment
      predictions. Segment polarity scores are obtained by assigning
      the document-level prediction to every segment. We can then
      produce finer-grained polarity distinctions via gating, 
      using the model's attention weights.
      %\item \textbf{\textsc{MilNet}:} Our multiple instance learning network.
\end{itemize}

\noindent We further illustrate the differences between
\textsc{HierNet} and \textsc{MilNet} in Figure \ref{fig:pipeline},
which includes short descriptions and simplified equations for each
model. \textsc{MilNet} naturally produces distinct segment polarities,
while \textsc{HierNet} assigns a single polarity score to every
segment. In both cases, gating is a further means of identifying
neutral segments.

Finally, we differentiate between variants of \textsc{HierNet} and \textsc{MilNet} according to:
\begin{itemize}[label={},itemindent=-1em,leftmargin=1em,itemsep=0em,topsep=1mm]
\item \textbf{Polarity source:} Controls whether we assign 
	polarities via \textit{segment-specific} or 
	\textit{document-wide} predictions. \textsc{HierNet} only 
	allows for document-wide predictions. \textsc{MilNet} can use both.
\item \textbf{Attention:} We use models without gating (no subscript),
	with gating (\textit{gt} subscript) as well as models trained with the
	attention mechanism disabled, falling back to simple averaging 
	(\textit{avg} subscript).
\end{itemize}

\subsection{Model Training and Evaluation}
\label{sec:train}

We trained \textsc{MilNet} and \textsc{HierNet} using Adadelta \cite{adadelta} 
for 25 epochs. Mini-batches of 200 documents 
were organized based on the reviews' segment and document lengths so 
the amount of padding was minimized. We used 300-dimensional 
\mbox{pre-trained} word2vec embeddings. We tuned hyper-parameters on the 
validation sets of the document classification collections, resulting in the 
following configuration (unless otherwise noted). For the CNN segment encoder, 
we used window sizes of 3, 4 and 5 words with 100 feature maps per window
size, resulting in 300-dimensional segment vectors. The GRU hidden
vector dimensions for each direction were set to 50 and the attention
vector dimensionality to 100. We used L2-normalization and dropout to
regularize the softmax classifiers and additional dropout on the
internal GRU connections. 

Real-valued polarity scores produced by the two models are mapped to
discrete labels using two appropriate thresholds \mbox{$t_1\,, t_2 \in
  [-1, 1]$}, so that a segment $s$ is classified as negative if
\mbox{$\pol(s) < t_1$}, positive if \mbox{$\pol(s) > t_2$} or neutral
otherwise.\footnote{The discretization of polarities is only used for
  evaluation purposes and is not necessary for summary extraction,
  where we only need a relative ranking of segments.} To evaluate
performance, we use macro-averaged F1 which is unaffected by class
imbalance.  We select optimal thresholds using 10-fold
cross-validation and report mean scores across folds.

The fully-supervised convolutional segment classifier \mbox{(Seg-CNN)} 
uses the same window size and
feature map configuration as our segment encoder. Seg-CNN was trained
on \textsc{SpoT} using segment labels directly and 10-fold
cross-validation (identical folds as in our main models). Seg-CNN is
not directly comparable to \textsc{MilNet} (or \textsc{HierNet}) due
to differences in supervision type (segment vs. document labels) and
training size (1K-2K segment labels vs. $\sim$250K document
labels). However, the comparison is indicative of the utility of fine-grained 
sentiment predictors that do not rely on expensive segment-level annotations.

\section{Results}

We evaluated models in two ways. We first assessed their ability to
classify segment polarity in reviews using the newly created
\textsc{SpoT} dataset and, additionally, the sentence corpora of
\newcite{kotzias2015group}. Our second suite of experiments focused on
opinion extraction: we conducted a judgment elicitation study to
determine whether extracts produced by \textsc{MilNet} are useful and
of higher quality compared to \textsc{HierNet} and other baselines. We
were also interested to find out whether EDUs provide a better basis
for opinion extraction than sentences.

\subsection{Segment Classification}
\label{sec:segclass}

\begin{table}[t]
\centering
\begin{tabular}{llcccc}
\toprule
& \multirow{2}{*}{\textbf{Method}}& \multicolumn{2}{c}{\textbf{Yelp'13$_{seg}$}} & \multicolumn{2}{c}{\textbf{IMDB$_{seg}$}} \\
& & Sent & EDU & Sent & EDU \\
\doubleRule
& Majority & 19.02$^\dagger\!\!$ & 17.03$^\dagger\!\!$ & 18.32$^\dagger\!\!$ & 21.52$^\dagger\!\!$ \\
%& Ablation & 39.81 & 42.88 & 50.56 & 43.84 \\
\midrule
\parbox[t]{.1mm}{\multirow{6}{*}{\rotatebox[origin=c]{90}{\textbf{Document}}}} & \textsc{HierNet$_{avg}\!\!\!$} & 54.21$^\dagger\!\!$ & 50.90$^\dagger\!\!$ & 46.99$^\dagger\!\!$ & 49.02$^\dagger\!\!$ \\
& \textsc{HierNet} & 55.33$^\dagger\!\!$ & 51.43$^\dagger\!\!$ & 48.47$^\dagger\!\!$ & 49.70$^\dagger\!\!$ \\
& \textsc{HierNet}$_{gt}$ & 56.64$^\dagger\!\!$ & 58.75 & 62.12 & 57.38$^\dagger\!\!$ \\
& \textsc{MilNet}$_{avg}$	& 58.43$^\dagger\!\!$ & 48.63$^\dagger\!\!$ & 53.40$^\dagger\!\!$ & 51.81$^\dagger\!\!$ \\
& \textsc{MilNet} & 52.73$^\dagger\!\!$ & 53.59$^\dagger\!\!$ & 48.75$^\dagger\!\!$ & 47.18$^\dagger\!\!$ \\
& \textsc{MilNet}$_{gt}$ & 59.74$^\dagger\!\!$ & 59.47 & 61.83$^\dagger\!\!$ & 58.24$^\dagger\!\!$ \\
\midrule
\parbox[t]{.1mm}{\multirow{3}{*}{\rotatebox[origin=c]{90}{\textbf{Segm}}}} & \textsc{MilNet}$_{avg}$ & 51.79$^\dagger\!\!$ & 46.77$^\dagger\!\!$ & 45.69$^\dagger\!\!$ & 38.37$^\dagger\!\!$ \\
& \textsc{MilNet} & 61.41 & 59.58 & 59.99$^\dagger\!\!$ & 57.71$^\dagger\!\!$ \\ 
& \textsc{MilNet}$_{gt}$ & \textbf{63.35} & \textbf{59.85} & \textbf{63.97} & \textbf{59.87}\\ 
\midrule
& SO-CAL & 56.53$^\dagger\!\!$ & 58.16$^\dagger\!\!$ & 53.21$^\dagger\!\!$ & 60.40 \\
& Seg-CNN & 56.18$^\dagger\!\!$ & 59.96 &  58.32$^\dagger\!\!$ &  62.95$^\dagger\!\!$ \\
\bottomrule
\end{tabular}
\vspace{-1ex}
\caption{Segment classification results (in macro-averaged F1). $\dagger$
  indicates that the system in question is significantly different
  from \textsc{MilNet}$_{gt}$ (approximate randomization test \protect\cite{Noreen:1989}, \mbox{$p< 0.05$}).}
\label{tbl:main}
\end{table}

Table~\ref{tbl:main} summarizes our results.  The first block in the
table reports the performance of the majority class baseline.
The second block considers models that do not utilize segment-level
predictions, namely \textsc{HierNet} which assigns polarity scores to
segments using its document-level predictions, as well as the variant
of \textsc{MilNet} which similarly uses document-level predictions
only (Equation \eqref{eq:sum}). In the third block, \textsc{MilNet}'s
segment-level predictions are used.  Each block further differentiates
between three levels of attention integration, as previously
described.  The final block shows the performance of SO-CAL and the
Seg-CNN classifier.

\begin{table}[t]
\footnotesize
\parbox{0.45\columnwidth}{
\begin{tabular}{@{~}l@{\hspace{1em}}l@{~}c@{~}c@{~}}
\toprule
\multicolumn{4}{c}{\textbf{Neutral Segments}}\\
& & Non-Gtd & Gated \\
\midrule
\parbox[t]{1mm}{\multirow{2}{*}{\rotatebox[origin=c]{90}{\textbf{Sent}}}} & \textsc{HierNet} & ~~4.67 & 36.60 \\
& \textsc{MilNet} & 39.61 & 44.60 \\
\bottomrule
& && \vspace{-0.5mm}\\
& & Non-Gtd & Gated \\
\midrule
\parbox[t]{1mm}{\multirow{2}{*}{\rotatebox[origin=c]{90}{\textbf{EDU}}}} & \textsc{HierNet} & ~~2.39 & 55.38 \\
& \textsc{MilNet} & 52.10 & 56.60 \\
\bottomrule
\end{tabular}
\vspace{-1ex}
\caption{F1 scores for neutral segments (Yelp'13).}
\label{tbl:neutral}
\centering
}
\hfill
\parbox{0.48\columnwidth}{
\begin{tabular}{lcc}
\toprule
\textbf{Method}& Yelp & IMDB \\
\doubleRule
GICF & 86.3 & 86.0 \\
GICF$_\textsc{HN}$ & 92.9 & 86.5 \\
%\textsc{MilNet}$_{avg}$+GICF & & 90.8 \\
GICF$_\textsc{MN}$ & 93.2 & 91.0 \\
\midrule
%\textsc{MilNet}$_{avg}$ & 93.1 & 90.6 \\
\textsc{MilNet} & 94.0 & 91.9 \\
\bottomrule
\end{tabular}
\vspace{.5mm}
\caption{Accuracy scores on the sentence classification datasets
  introduced in \protect\newcite{kotzias2015group}.}
\label{tbl:kotzias}
}
\end{table}

When considering models that use document-level supervision,
\textsc{MilNet} with gated, segment-specific polarities 
obtains the best classification performance across all four datasets.
Interestingly, it performs comparably to Seg-CNN, the fully-supervised
segment classifier, which provides additional evidence that
\textsc{MilNet} can effectively identify segment polarity without the
need for segment-level annotations.  Our model also outperforms the
strong SO-CAL baseline in all but one datasets which is remarkable
given the expert knowledge and linguistic information used to develop
the latter.  Document-level polarity predictions result in lower
classification performance across the board. Differences between the
standard hierarchical and multiple instance networks are less
pronounced in this case, as \textsc{MilNet} loses the advantage of
producing segment-specific sentiment predictions. Models without
attention perform worse in most cases.  The use of gated
polarities benefits all model configurations, indicating the method's
ability to selectively focus on segments with significant sentiment cues.

\begin{figure}[t!]
	\centering
	\begin{minipage}{\columnwidth}
	\includegraphics[width=\columnwidth]{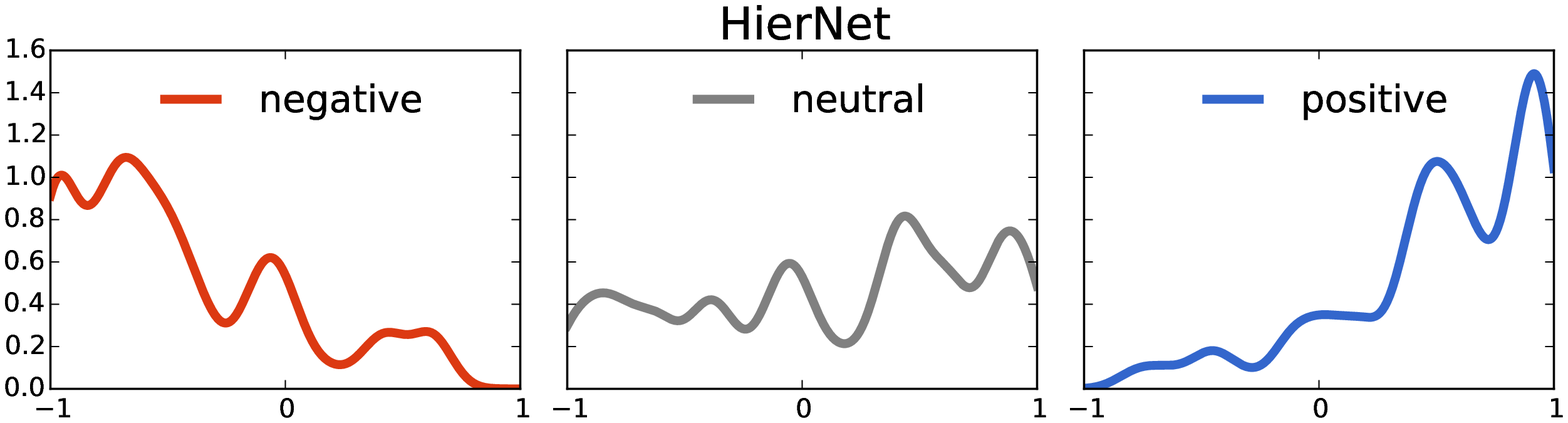}
	\end{minipage}\vspace{-2mm}
	\begin{minipage}{\columnwidth}
	\includegraphics[width=\columnwidth]{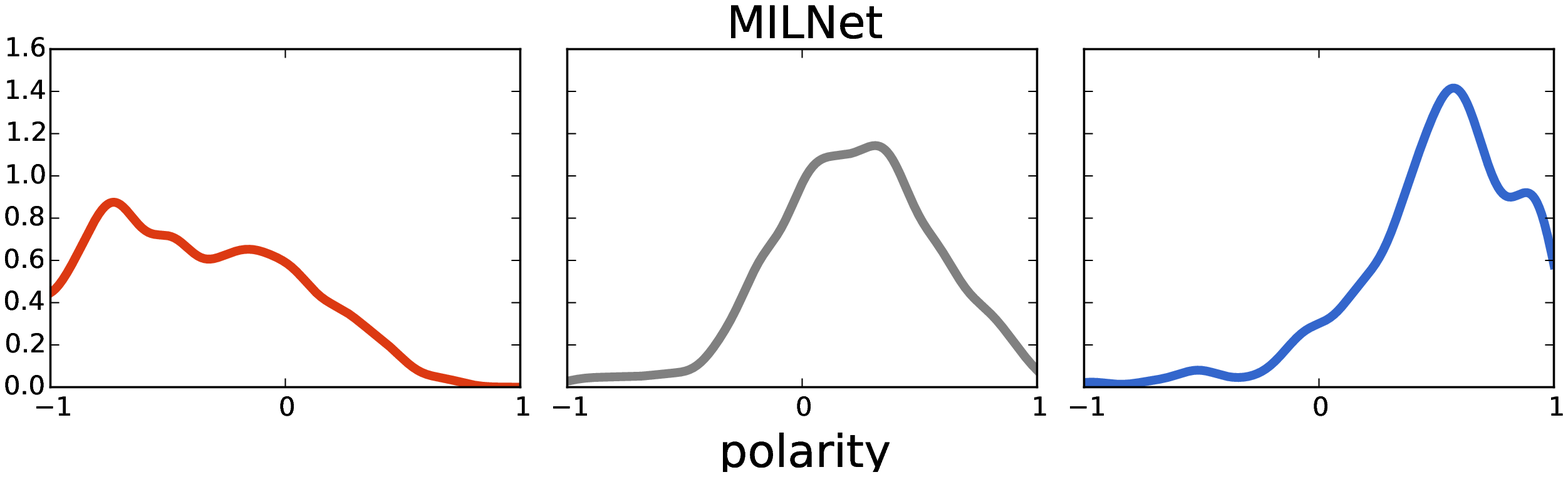}
	\end{minipage}
\vspace{-1ex}
	\caption{Distribution of predicted polarity scores across
          three classes (Yelp'13 sentences).}
	\label{fig:kde}
\end{figure}

We further analyzed the polarities assigned by \textsc{MilNet} and
\textsc{HierNet} to positive, negative, and neutral segments.
Figure~\ref{fig:kde} illustrates the distribution of polarity scores
produced by the two models on the Yelp'13 dataset (sentence
segmentation). In the case of negative and positive sentences, both
models demonstrate appropriately skewed distributions. However, the
neutral class appears to be particularly problematic for
\textsc{HierNet}, where polarity scores are scattered across a wide
range of values. In contrast, \textsc{MilNet} is more successful at
identifying neutral sentences, as its corresponding distribution has a
single mode near zero. Attention gating addresses this issue by moving
the polarity scores of sentiment-neutral segments towards zero. This
is illustrated in Table~\ref{tbl:neutral} where we observe that gated
variants of both models do a better job at identifying neutral
segments. The effect is very significant for \textsc{HierNet}, while
\textsc{MilNet} benefits slightly and remains more effective
overall. Similar trends were observed in all four \textsc{SpoT}
datasets.

\begin{figure}[t]
\includegraphics[width=\columnwidth]{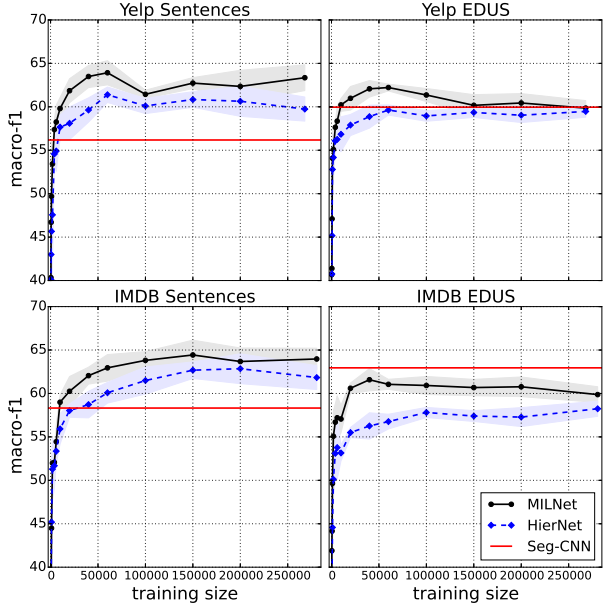}
\caption{Performance of \textsc{HierNet}$_{gt}$ and \mbox{\textsc{MilNet}$_{gt}$} for varying 
training sizes.}
\label{fig:lcurve}
\end{figure}

In order to examine the effect of training size, we trained multiple
models using subsets of the original document collections. We trained
on five random subsets for each training size, ranging from 100
documents to the full training set, and tested segment classification
performance on \textsc{SpoT}. The results, averaged across trials, are
presented in Figure~\ref{fig:lcurve}. With the exception of the IMDB
EDU-segmented dataset, \textsc{MilNet} only requires a few thousand
training documents to outperform the supervised Seg-CNN.
\textsc{HierNet} follows a similar curve, but is inferior to
\textsc{MilNet}. A reason for \textsc{MilNet}'s inferior performance
on the IMDB corpus (EDU-split) can be low-quality EDUs, due to the
noisy and informal style of language used in IMDB reviews.

Finally, we compared \textsc{MilNet} against the GICF model
\cite{kotzias2015group} on their Yelp and IMDB sentence sentiment
datasets.\footnote{GICF only handles binary labels, which makes it
  unsuitable for the full-scale comparisons in
  Table~\ref{tbl:main}. Here, we binarize our training datasets and
  use same-sized sentence embeddings for all four models
  ($\mathbb{R}^{150}$ for Yelp, $\mathbb{R}^{72}$ for IMDB).}  Their
model requires sentence embeddings from a pre-trained neural model.
We used the hierarchical CNN from their work
\cite{denil2014extraction} and, additionally, pre-trained
\textsc{HierNet} and \textsc{MilNet} sentence embeddings. The results
in Table~\ref{tbl:kotzias} show that \textsc{MilNet} outperforms all
variants of GIFC. Our models also seem to learn better sentence
embeddings, as they improve GICF's performance on both collections.

\begin{table}[t]
\small
\centering
\begin{tabular}{@{~}lc@{~}c@{~}c@{~}}
\toprule
\textbf{Method} & \textbf{Informativeness}& \textbf{Polarity} & \textbf{Coherence} \\
\doubleRule
\textsc{HierNet}$^{sent}$ & 43.7 & 33.6 & 43.5\\
\textsc{MilNet}$^{sent}$ & \textbf{45.7} & \textbf{36.7} & \textbf{44.6}\\
Unsure & 10.7 & 29.6 & 11.8\\
\midrule
\textsc{HierNet}$^{edu}$ & 34.2$^\dagger\!\!$ & 28.0$^\dagger\!\!$ & \textbf{48.4}\\
\textsc{MilNet}$^{edu}$ & \textbf{53.3} & \textbf{61.1} & 45.0\\
Unsure & 12.5 & 11.0 & ~~6.6\\
\midrule
\textsc{MilNet}$^{sent}$ & 35.7$^\dagger\!\!$ & 33.4$^\dagger\!\!$ & \textbf{70.4}$^\dagger\!\!$\\
\textsc{MilNet}$^{edu}$ & \textbf{55.0} & \textbf{51.5} & 23.7\\
Unsure & ~~9.3 & 15.2 & ~~5.9\\
\midrule
\textsc{Lead} & 34.0 & 19.0$^\dagger\!\!$ & \textbf{40.3}\\
\textsc{Random} & 22.9$^\dagger\!\!$ & 19.6$^\dagger\!\!$ & 17.8$^\dagger\!\!$\\
\textsc{MilNet}$^{edu}$ & \textbf{37.4} & \textbf{46.9} & 33.3\\
Unsure & ~~5.7 & 14.6 & ~~8.6\\
\bottomrule
\end{tabular}
\vspace{-1ex}
\caption{Human evaluation results (in percentages). $\dagger$
  indicates that the system in question is significantly different
  from \textsc{MilNet} (sign-test, \mbox{$p< 0.01$}).}
\label{tbl:human}
\end{table}

\begin{figure*}[t]
\footnotesize
\centering
\begin{mdframed}
[frametitle={\footnotesize \textnormal{[Rating: {\Large \mystar \mystar \mystar \mystar}] As with any family-run hole in the wall, service can be slow. What the staff lacked in speed, they made up for in charm. The food was good, but nothing wowed me. I had the Pierogis while my friend had swedish meatballs. Both dishes were tasty, as were the sides. One thing that was disappointing was that the food was a a little cold (lukewarm). The restaurant itself is bright and clean. I will go back again when i feel like eating outside the box.}}]

\hspace{-1mm}\rotatebox[origin=c]{90}{EDU-based~~}\vspace{-1mm}{
\adjustbox{minipage=\columnwidth,margin=0em,width=\columnwidth,center,bgcolor=light-gray}{%
  ~~~Extracted via \textsc{HierNet}$_{gt}$\vspace{1mm}\\
  \textcolor{pos}{~~~\textcolor{black}{(0.13)} [+0.26] The food was good$^+$}\\
  \textcolor{pos}{~~~\textcolor{black}{(0.10)} [+0.26] but nothing wowed me.$^+$}\\
  \textcolor{pos}{~~~\textcolor{black}{(0.09)} [+0.26] The restaurant itself is bright and clean$^+\!\!$}\\
  \textcolor{pos}{~~~\textcolor{black}{(0.13)} [+0.26] Both dishes were tasty$^+$}\\
  \textcolor{pos}{~~~\textcolor{black}{(0.18)} [+0.26] I will go back again$^+$}\\
}}%
{\adjustbox{minipage=\columnwidth,margin=0em,width=\columnwidth,center=-3mm,bgcolor=light-gray}{%
  ~~~Extracted via \textsc{MilNet}$_{gt}$\vspace{1mm}\\
  \textcolor{pos}{~~~\textcolor{black}{(0.16)} [+0.12] The food was good$^+$}\\
  \textcolor{pos}{~~~\textcolor{black}{(0.12)} [+0.43] The restaurant itself is bright and clean$^+\!\!$}\\
  \textcolor{pos}{~~~\textcolor{black}{(0.19)} [+0.15] I will go back again$^+$}\\
  \textcolor{neg}{~~~\textcolor{black}{(0.09)} [--0.07] but nothing wowed me.$^-$}\\
  \textcolor{neg}{~~~\textcolor{black}{(0.10)} [--0.10] the food was a a little cold (lukewarm)$^-$}\\
}}

\vspace{-1mm}\hrule\vspace{2mm}
\hspace{-1mm}\rotatebox[origin=c]{90}{Sent-based}{
\adjustbox{minipage=\columnwidth,margin=0em,width=\columnwidth,center,bgcolor=light-gray}{%
  \textcolor{pos}{~~~\textcolor{black}{(0.12)} [+0.23] Both dishes were tasty, as were the sides$^+$}\\
  \textcolor{pos}{~~~\textcolor{black}{(0.18)} [+0.23] The food was good, but nothing wowed me$^+\!\!\!$}\\
  \textcolor{pos}{~~~\textcolor{black}{(0.22)} [+0.23] One thing that was disappointing was that}\\
  \textcolor{pos}{~~~~~~~~~~~~~~~~~~~~~~~~~~$\;$the food was a a little cold (lukewarm)$^+$}%
}}%
{\adjustbox{minipage=\columnwidth,margin=0em,width=\columnwidth,center=-3mm,bgcolor=light-gray}{%
  \textcolor{pos}{~~~\textcolor{black}{(0.13)} [+0.26] Both dishes were tasty, as were the sides$^+$}\\
  \textcolor{pos}{~~~\textcolor{black}{(0.20)} [+0.59] I will go back again when I feel like eating}\\
  \textcolor{pos}{~~~~~~~~~~~~~~~~~~~~~~~~~~$\;$outside the box$^+$}\\
  \textcolor{neg}{~~~\textcolor{black}{(0.18)} [--0.12] The food was good, but nothing wowed me$^-\!\!\!$}
}}

\vspace{1mm}\hrule\vspace{2mm}
{\adjustbox{minipage=\columnwidth,margin=2.5em 0em,width=\columnwidth,center=143mm,bgcolor=light-gray}{%
\mbox{(number): attention weight~~~~~[number]: non-gated polarity score~~~~~\textcolor{pos}{text$^+$}: extracted positive opinion~~~~~\textcolor{neg}{text$^-$}: extracted negative opinion}
}}
\end{mdframed}
\caption{Example EDU- and sentence-based opinion summaries produced by 
\textsc{HierNet}$_{gt}$ and \textsc{MilNet}$_{gt}$.}
\vspace{-1mm}
\label{fig:summ-comp}
\end{figure*}

\subsection{Opinion Extraction}
\label{sec:human}

In our opinion extraction experiments, AMT workers (all native English
speakers) were shown an original review and a set of extractive, bullet-style
summaries, produced by competing systems using a 30\% compression
rate. Participants were asked to decide which summary was best
according to three criteria: \textit{Informativeness} (Which summary
best captures the salient points of the review?), \textit{Polarity}
(Which summary best highlights positive and negative comments?)  and
\textit{Coherence} (Which summary is more coherent and easier to
read?). Subjects were allowed to answer ``Unsure'' in cases where they
could not discriminate between summaries.  We used all reviews
from our \textsc{SpoT} dataset and collected three responses per
document. We ran four judgment elicitation studies: one comparing
\textsc{HierNet} and \textsc{MilNet} when summarizing reviews
segmented as sentences, a second one comparing the two models with EDU
segmentation, a third which compares EDU- and sentence-based summaries
produced by \textsc{MilNet}, and a fourth where EDU-based summaries
from \textsc{MilNet} were compared to a \textsc{Lead} (the first $N$
words from each document) and a \textsc{Random} (random EDUs)
baseline.

Table~\ref{tbl:human} summarizes our results, showing the proportion
of participants that preferred each system. The first block in the
table shows a slight preference for \textsc{MilNet} across criteria. 
The second block shows significant preference for \textsc{MilNet} against 
\textsc{HierNet} on informativeness and polarity, whereas \textsc{HierNet} was 
more often preferred in terms of coherence, although the difference is not
statistically significant. The third block compares sentence and EDU
summaries produced by \textsc{MilNet}. EDU summaries were perceived as
significantly better in terms of informativeness and polarity, but not
coherence. This is somewhat expected as EDUs tend to produce more
terse and telegraphic text and may seem unnatural due to segmentation
errors. In the fourth block we observe that participants find
\textsc{MilNet} more informative and better at distilling polarity
compared to the \textsc{Lead} and \textsc{Random} (EDUs) baselines. We
should point out that the \textsc{Lead} system is not a strawman;
it has proved hard to outperform by more sophisticated methods
\cite{Nenkova:2005}, particularly on the newswire domain.

Example EDU- and sentence-based summaries produced by gated variants of
\textsc{HierNet} and \textsc{MilNet} are shown in Figure~\ref{fig:summ-comp},
with attention weights and polarity scores of the extracted segments shown
in round and square brackets respectively. For both granularities, 
\textsc{HierNet}'s positive document-level prediction results in a single 
polarity score assigned to every segment, and further adjusted using
the corresponding attention weights. The extracted segments are
informative, but fail to capture the negative sentiment of some segments. 
In contrast, \textsc{MilNet} is able to detect positive and negative
snippets via individual segment polarities. Here, EDU segmentation produced a 
more concise summary with a clearer grouping of positive and negative snippets.

\section{Conclusions}
\label{sec:conc}

In this work, we presented a neural network model for fine-grained
sentiment analysis within the framework of multiple instance
learning. Our model can be trained on large scale sentiment
classification datasets, without the need for segment-level labels.
As a departure from the commonly used vector-based composition, our
model first predicts sentiment at the sentence- or EDU-level and
subsequently combines predictions up the document hierarchy. An
attention-weighted polarity scoring technique provides a natural way
to extract sentiment-heavy opinions. Experimental results demonstrate
the superior performance of our model against more conventional neural
architectures. Human evaluation studies also show that \textsc{MilNet}
opinion extracts are preferred by participants and are effective at
capturing informativeness and polarity, especially when using EDU
segments. In the future, we would like to focus on multi-document, 
aspect-based extraction \cite{cao2017improving} and ways of improving the
coherence of our summaries by taking into account more fine-grained
discourse information \cite{marcu2002}.

\section*{Acknowledgments} 
The authors gratefully acknowledge the support of the European
Research Council (award number 681760). We thank TACL action editor 
Ani Nenkova and the anonymous reviewers whose feedback helped improve the present paper, 
as well as Charles Sutton, Timothy Hospedales, and members of EdinburghNLP for
helpful discussions and suggestions.

\bibliography{milnet-sent-summ}
\bibliographystyle{acl2012}

\end{document}